\documentclass[10pt,journal,compsoc]{IEEEtran}
\usepackage{footnote}
\usepackage{graphicx}

\usepackage[table]{xcolor}
\usepackage{multirow}
\usepackage{hhline}
\usepackage{rotating}
\usepackage{fixltx2e}
\usepackage{eurosym} 

\ifCLASSOPTIONcompsoc
  \usepackage[nocompress]{cite}
\else
  \usepackage{cite}
\fi
\ifCLASSINFOpdf
\else
\fi
\usepackage{amsmath}
\usepackage{algorithmic}

\usepackage{array}

\ifCLASSOPTIONcompsoc
  \usepackage[caption=false,font=footnotesize,labelfont=sf,textfont=sf]{subfig}
\else
  \usepackage[caption=false,font=footnotesize]{subfig}
\fi
\usepackage{fixltx2e}

\usepackage[bottom]{footmisc}
\usepackage{stfloats}
\begin{document}
%
\title{Artificial neural networks in action for an automated cell-type classification of biological neural networks}

\author{\IEEEauthorblockN{Eirini Troullinou\IEEEauthorrefmark{1}\IEEEauthorrefmark{2},
Grigorios Tsagkatakis\IEEEauthorrefmark{2},
Spyridon Chavlis\IEEEauthorrefmark{2}\IEEEauthorrefmark{3},\\
Gergely Turi\IEEEauthorrefmark{4},
Wen-Ke Li\IEEEauthorrefmark{4},
Attila Losonczy\IEEEauthorrefmark{4},
Panagiotis Tsakalides\IEEEauthorrefmark{1}\IEEEauthorrefmark{2}, and Panayiota Poirazi\IEEEauthorrefmark{3}}\\
\IEEEauthorblockA{\IEEEauthorrefmark{1}Department of Computer Science, University of Crete, Heraklion, 70013, Greece}\\
\IEEEauthorblockA{\IEEEauthorrefmark{2}Institute of Computer Science, Foundation for Research and Technology Hellas, Heraklion, 70013, Greece}\\
\IEEEauthorblockA{\IEEEauthorrefmark{3}Institute of Molecular Biology and Biotechnology, Foundation for Research and Technology Hellas, Heraklion, 70013, Greece}\\
\IEEEauthorblockA{\IEEEauthorrefmark{4}Department of Neuroscience, Columbia University Medical Center, New York, USA}}

%
%



\markboth{Accepted at IEEE Transactions on Emerging Topics in Computational Intelligence-2019-0236}%
{Shell \MakeLowercase{\textit{et al.}}: Bare Demo of IEEEtran.cls for Computer Society Journals}

%



\IEEEtitleabstractindextext{%
\begin{abstract}
Identification of different neuronal cell types is critical for understanding their contribution to brain functions. Yet, automated and reliable classification of neurons remains a challenge, primarily because of their biological complexity. Typical approaches include laborious and expensive immunohistochemical analysis while feature extraction algorithms based on cellular characteristics have recently been proposed. The former rely on molecular markers, which are often expressed in many cell types, while the latter suffer from similar issues: finding features that are distinctive for each class has proven to be equally challenging. Moreover, both approaches are time consuming and demand a lot of human intervention. In this work we establish the first, automated cell-type classification method that relies on neuronal activity rather than molecular or cellular features. We test our method on a real-world dataset comprising of raw calcium activity signals for four neuronal types. We compare the performance of three different deep learning models and demonstrate that our method can achieve automated classification of neuronal cell types with unprecedented accuracy.
\end{abstract}

}

\maketitle

\IEEEdisplaynontitleabstractindextext

%
\IEEEpeerreviewmaketitle

\IEEEraisesectionheading{\section{Introduction}\label{sec:introduction}}

\newcommand\blfootnote[1]{%
  \begingroup
  \renewcommand\thefootnote{}\footnote{#1}%
  \addtocounter{footnote}{-1}%
  \endgroup
}

\blfootnote{\textsuperscript{\textcopyright} 20XX IEEE.  Personal use of this material is permitted.  Permission from IEEE must be obtained for all other uses, in any current or future media, including reprinting/republishing this material for advertising or promotional purposes, creating new collective works, for resale or redistribution to servers or lists, or reuse of any copyrighted component of this work in other works.}

%
%
%
%
\IEEEPARstart{T}{o} understand the function -and dysfunction- of neural circuits we must first identify the subtypes of neurons that comprise these networks, their biophysical and anatomical characteristics as well as their inter-dependencies. Different cell types typically exhibit different anatomy, connectivity and/or biophysical properties which, in turn, influence their specific function and role in pathologies such as epilepsy \cite{trevelyan2006modular,freund2007perisomatic}, anxiety disorder \cite{freund2007perisomatic,powell2003genetic}, the Tourette syndrome \cite{kalanithi2005altered}, autism \cite{tabuchi2007neuroligin}, the Rett syndrome \cite{dani2005reduced} and schizophrenia \cite{gonzalez2008gaba,lewis2005cortical}. The development of a cellular taxonomy will thus facilitate our understanding of both healthy and diseased brain functioning, as most brain pathologies affect specific neuronal types \cite{muratore2017cell,skene2016identification}. 

Despite this pressing need, neuronal classification remains challenging. Traditional approaches rely on qualitative descriptors, such as the expression of specific molecular markers (proteins) in combination with their anatomical characteristics and laminar localization \cite{zeng2017neuronal, maccaferri2003interneuron,booker2018morphological}. These approaches require sacrificing the animal and slicing its brain, so as to image cells under a confocal microscope and measure the expression levels of specific markers. Moreover, during this process, many cells die or cannot be identified, requiring that the process is repeated over many animals. 

Given the magnitude and complexity of neuronal classification, the time-consuming and expensive experiments that are required, and because the manual classification attempts using qualitative descriptors are ill-equipped to deal with big data, high-throughput technologies are demanded. Hence, as discussed in the Related Work section, quantitative methods have recently been developed. These methods are based on morphological, physiological, molecular, and/or electrophysiological characteristics of neurons \cite{guerra2011comparison,vasques2016morphological, ascoli2008petilla, mcgarry2010quantitative} and use supervised or unsupervised classifiers, providing a quantitative and unbiased identification of distinct subtypes when applied to selected datasets. However, obtaining such characteristic features for different cell classes is still a challenging task, primarily because the characteristic features of specific neurons and their uniqueness remain largely unknown \cite{zeng2017neuronal}. Thus, feature extraction-based algorithmic approaches entail laborious and expensive experimentation, often involving several different techniques, which limits the attractiveness of automated classifiers relying on such features.

In this work, we introduce a novel method for automated cell-type recognition that relies on a feature never exploited before: the timeseries of calcium (Ca\textsuperscript{2+}) activity signals of neurons, as measured with imaging techniques in the behaving animal. Specifically, we use timeseries describing the activity of four neuronal cell-types in the CA1 subregion of the hippocampus: excitatory pyramidal cells (PY) and three GABAergic interneuronal subtypes, namely parvalbumin-positive (PV), somatostatin-positive (SOM) and vasoactive intestinal polypeptide-positive (VIP) cells. Neuronal activity is measured using Ca\textsuperscript{2+} imaging, which is a powerful technique for monitoring the activity of distinct neurons in brain tissue \textit{in vivo} \cite{stosiek2003vivo} and is currently the most popular recording technique for behaving animals \cite{birkner2017improved,resendez2015vivo, gobel2007vivo}. Here, motivated by the challenges that were discussed previously, we examine the potential of replacing existing approaches (i.e., feature extraction-based algorithms as well as immunohistochemical analysis methods) with fast, reliable, cell-type classification, which is based on Ca\textsuperscript{2+} imaging recordings using state-of-the-art Deep Learning (DL) architectures for timeseries analysis. Towards this goal, we consider the raw fluorescence signal without any preprocessing. To our knowledge, this is the first successful attempt to classify cell-types based solely on the raw Ca\textsuperscript{2+} activity signal.


With the advent of DL, various approaches and models for timeseries analysis and forecast have been developed \cite{ravanelli2018light,hou2018audio, zheng2017video}. For different fields and applications, suitable algorithms and models vary depending on the nature and purpose of the data. For the task of neuronal cell-type classification considered in this work, we employ three types of network architectures, namely 1-Dimensional Convolutional Neural Networks (1D-CNNs), Recurrent Neural Networks (RNNs) and Long Short-Term Memory Networks (LSTMs), which are widely used in timeseries analysis. Our motivation for applying these specific 3 models is that each of these frameworks is used with particular types of datasets, as the 1D-CNN model can exploit temporal locality in timeseries data, while the other two architectures and especially the LSTM model are ideal for revealing long-term dependencies. Given that this is the first work on neuronal cell-type classification using DL models with Ca\textsuperscript{2+} signal data, no prior knowledge exists on how Ca\textsuperscript{2+} signal timeseries are best modelled, so that neuronal cell-types can be most accurately inferred. Thus, we conducted a comparative analysis, in order to examine which of them is the most appropriate for this specific task. 

RNN and LSTM are popular models for timeseries data, we thus also included them in our analysis. RNNs and LSTMs are known to require a serious amount of training time when processing long timeseries data and several studies have focused on the acceleration of these models via complex algorithmic frameworks \cite{khomenko2016accelerating, kuchaiev2017factorization, neil2016phased}. Here, as discussed in the methodology subsection, we propose a simple scheme, i.e. a data re-organization, which significantly reduces the number of parameters, thus substantially accelerating the training of these networks when using long timeseries data. 


In summary, we present a novel deep learning approach for the automated classification of neuronal cell types that relies on a feature never used before, namely the calcium activity signal of neurons in the behaving animal. We develop and compare different deep learning models and demonstrate their ability to achieve high accuracy on a real-world dataset with four neuronal types. This is the first demonstration that automated neuronal classification can be achieved reliably, without requiring laborious, expensive and highly invasive experiments. 

The remainder of the paper is organized as follows: In section II, we report the motivation and contribution of our study as well as the related, prior work on neuronal cell type classification. In Section III, we describe and analyze the proposed approaches. Experimental results are presented in Section IV and conclusions are drawn in Section V.  



 




\section{Objectives, Related Work and Contribution}
\noindent
Our study investigates whether the selected DL methods (i.e., 1D-CNN, RNN and LSTM) can solve the problem of neuronal cell-type classification when utilizing as input a single signal type, namely the raw Ca\textsuperscript{2+} activity signal of neurons recorded in the CA1 area of the hippocampus from behaving mice. We focus on a four-class problem, consisting of PY cells and three GABAergic interneurons (IN) subtypes (PV, SOM and VIP cells).

\subsection{Related Work}

Timeseries analysis using DL architectures is among the most active areas of research in machine learning. DL architectures like CNN, RNN and LSTM have been used for a multitude of different tasks including the analysis of measurements from wearable sensors , modeling networks traffic for cybersecurity, and high frequency trading in finance among others \cite{fawaz2019deep}. For the case of biological data analysis, application of DL and traditional machine learning methods are presented next.

Cell-type classification has been extensively studied in the past decades, however it was primarily aimed at discriminating between healthy and malignant cells \cite{su2014neural, tomari2014computer, geetha2019cervical, shah2013comparison}. Hence, little progress has been made in the automated classification of neuronal cell-types. For example, it is widely known that the two major neuronal types of the mammalian brain are the pyramidal cells and the  GABAergic IN. While both PY and IN vary substantially in their anatomy and biophysical properties across brain areas, a clear classification of the different cell types has yet to be achieved. INs in particular come in many different types and shapes \cite{kepecs2014interneuron}, even within the same brain region. Specifically, the hippocampus has over 20 different types of INs \cite{pelkey2017hippocampal}. In this section we review several qualitative as well as quantitative approaches, whose ultimate goal is to automate neuronal cell-type classification. 

To aid ongoing efforts towards IN classification, to facilitate the exchange of information and to build a foundation for future progress in the field, the Petilla Interneuron Nomenclature Group (PING), proposed a standardized nomenclature of IN properties \cite{ascoli2008petilla} by defining qualitative descriptors (i.e., key features) that can be used for their identification. Such features are morphological, molecular, physiological and biophysical properties of these cells. Based on these qualitative descriptors, Zeng et al' \cite{zeng2017neuronal} reviewed high-throughput classification methods, such as light microscopy, electron microscopy, optical imaging of electrical activity and molecular profiling, which enable the collection of morphological, physiological and molecular data from large numbers of neurons.

Guerra et al' \cite{guerra2011comparison} explored the utilization of supervised and unsupervised classification algorithms to distinguish INs from PY, based solely on their morphological features. They used a database of 128 PY and 199 INs from mouse neocortex, and for each cell, 65 morphological features were measured, creating a data matrix. Their main finding was that supervised classification methods outperformed unsupervised algorithms (hierarchical clustering), and thus the latter approach is not as effective as supervised classification when distinguishing between the aforementioned cell types. Eventually, they showed that the selection of subsets of distinguishing features enhanced
the classification accuracy for both sets of algorithms.

Vasques et al' \cite{vasques2016morphological} used 43 morphological features as predictors and showed the results of applying supervised and unsupervised classification techniques in order to distinguish neuronal cell-types. More specifically, they assessed and compared the accuracy of different classification algorithms trained on 430 digitally reconstructed neurons and classified them according to layer and/or m-type (i.e morphology) with young and/or adult developmental state. Their findings regarding the superiority of supervised algorithms against unsupervised coincide with those of study \cite{guerra2011comparison}.

DeFelipe et al' \cite{defelipe2013new} proposed a classification scheme based on 6 axonal features, which are considered as a representative subset of axonal morphological properties that could be suitable for IN classification. They also designed and deployed an interactive web-based system to empirically test the level of agreement among 42 experts in assigning the 6 features to individual cortical INs. Specifically, experienced neuroscientists were asked to ascribe the categories they considered most appropriate to each neuron (there were 6 features and 21 categories in total, and thus, based on each feature, the neuron would be ascribed to one out of the 2 or more categories that corresponded to the specific feature). For some of the proposed features, there were high levels of observed agreement between experts in the classification of neurons, while for other features there was a low level of inter-expert agreement. The ultimate goal of their experiment was to build a model that could classify a neuron on the basis of its morphological characteristics and more specifically, in terms of the 6 features defined in their study. 

Overall, these studies highlight the current state-of-the-art in cell-type classification in a feature-based manner. However, to our knowledge, there is no prior work on neuronal cell-type classification that relies solely on the Ca\textsuperscript{2+} activity of neurons recorded \textit{in vivo} from behaving animals. As such, a major novelty of this work is the application of DL-based time-series analysis methods in cell classification.

\subsection{Our contribution}
In this work, we employ and compare the performance of 1D-CNN, RNN, and LSTM models on the task of neuronal cell-type classification based solely on the raw Ca\textsuperscript{2+} signals measured in different types of neurons while mice perform a goal oriented task \cite{turi2019vasoactive}. A comparative research analysis on the specific application using the aforementioned models is missing from the existing literature, and to the best of our knowledge, this is the first time that algorithmic models use raw Ca\textsuperscript{2+} signals to classify different neuronal types \textit{in vivo}. As discussed in the Related Work section, most of the algorithmic approaches use morphological features of neurons, with numerous limitations: (1) the geometry of individual neurons varies significantly within the same class, (2) different techniques are used to extract morphologies (e.g., imaging, histology, and reconstruction techniques) and (3) there is high inter-laboratory variability, all of which introduce substantial variability in the measured characteristics \cite{scorcioni2004quantitative}. 

\par Moreover, RNN and LSTM models require a considerable amount of time when trained with long timeseries data \cite{khomenko2016accelerating, kuchaiev2017factorization, neil2016phased}. Thus, in our work as analytically discussed in the Methodology subsection, we propose a simple scheme, i.e. a data re-organization, which significantly reduces the number of parameters, thus accelerating the training of these networks when used with long timeseries sequences.

Our work utilizes Ca\textsuperscript{2+} imaging, which is currently the most widely used technique for recording the activity of neurons in the behaving animal. It allows the simultaneous recording of tens to hundreds of cells without causing any damage to the neural tissue of interest, as opposed to more invasive, electrode-based recording methods (e.g., tetrodes, octrodes and silicon probes). Compared to these methods, Ca\textsuperscript{2+} imaging is also more stable, as the same neurons can be recorded over time periods that extend from days to months. Nevertheless, due to its slow kinetics, the signal produced is not ideal for resolving single spikes and makes the inference of neuronal cell-types non-trivial. Thus, cell-type classification relies mostly on the expression of specific molecular markers \cite{zeng2017neuronal}, which however are limited to just a small number of classes, many of which comprise of several sub-types of INs. Moreover, such an approach is laborious, expensive and requires the use of many different transgenic animal lines.

To address these limitations, our work makes the following contributions to the problem of automatic neuronal cell-type classification: 
\begin{itemize}
  \item We present a comparative research analysis of the 1D-CNN, RNN and LSTM models, in the domain of timeseries analysis for the task of neuronal cell-type classification, where such an analysis is missing from the existing literature.
  \item To the best of our knowledge, this is the first time that algorithmic models use raw Ca\textsuperscript{2+} signals to classify different neuronal types. The models are based solely on the Ca\textsuperscript{2+} imaging signatures of the neuronal types, unlike existing post-hoc techniques.
  \item We propose a simple and easily applicable scheme, which accelerates the training time of RNN and LSTM models when used with long timeseries data.
  \item  Our proposed approach (i.e., DL models based only on Ca\textsuperscript{2+} signals) is feature-independent and can thus replace feature extraction-based methods. 
  \item Performance accuracy is very high, suggesting that a gradual substitution of immunohistochemical analysis could be potentially considered, as such analyses are laborious, time-consuming and very expensive. 
\end{itemize}

\section{Proposed Approaches}
\noindent
To assess whether four cell types (PY, PV, SOM and VIP), whose activity is described with timeseries of raw Ca\textsuperscript{2+} imaging data can be correctly classified, we employ the following classification models and compare their performance:

\begin{itemize}
  \item 1-Dimensional Convolutional Neural Networks (1D-CNN) 
  \item Recurrent Neural Networks (RNN)
  \item Long Short-Term Memory Networks (LSTM) 
\end{itemize}

\subsection{Preliminary Concepts}
\noindent
In this subsection we discuss the theoretical background of the proposed approaches.

\subsubsection{Convolutional Neural Networks}
\noindent
A Convolutional Neural Network (CNN) is a class of deep neural networks (DNN), most commonly applied to imaging applications that consists of input, output and hidden layers of nodes along with their respective connections that encode the learnable weights of the network. CNNs are regularized versions of traditional Multilayer Perceptrons (MLPs), which usually refer to fully connected networks, i.e., each node in one layer connects to all nodes in the next layer. This characteristic of MLPs makes them prone to overfitting during training. Hence, one of the distinguishing property of CNNs compared to MLPs is the local connectivity among nodes. When dealing with high-dimensional inputs such as Ca\textsuperscript{2+} imaging timeseries data, it is impractical to connect nodes in one layer with all nodes in the previous volume because such a network architecture does not take the overall structure of the data into account. 

CNNs exploit local correlations by enforcing a sparse local connectivity pattern between neurons of adjacent layers, i.e., each node is connected to only a small region of the input signal. Namely, in a convolutional layer, nodes receive input from only a restricted subarea of the previous layer and this input area of a node is called its receptive field. Another distinguishing feature of CNNs is the shared weights. In CNNs, each filter is replicated across the entire receptive field. These replicated units share the same parameterization (weight vector and bias) and form a feature map, i.e., all nodes in a given convolutional layer respond to the same feature within their specific receptive field. Replicating units in this way allows for features to be detected regardless of their position in the visual field, thus constituting a property of translation invariance. 

\subsubsection{Typical Architecture of a 1-Dimensional CNN}
~\newline
A typical 1D-CNN architecture, as shown in Fig. \ref{fig_architectures}(c) is formed by a stack of distinct layers that transform the input volume into an output volume (holding the class scores) through a differentiable function. The core building block of a CNN is the convolutional layer, whose parameters' consist of a set of learnable filters (or kernels), which have a small receptive field. Given an input vector $x \in R^{1\times N}$ and a trainable filter $f \in R^{1\times K}$, the convolution of the two entities results in an output vector $c \in R^{1\times M}$, where $M=N-K+1$. The value of $M$ may vary based on the stride of the operation of convolution, with bigger strides leading to smaller outputs. 

The trainable parameters of the network, i.e., the filter and the bias are initialized randomly, but as the network is trained using the backpropagation learning algorithm, they are optimized and are able to capture important features from the given inputs. In order to construct a reliable network that will be able to capture complex and abstract features from the input data, we need to build a deep architecture comprised with more than one convolutional layer. As we go through layers in deep architectures, the features not only increase in number (depth size) but also in complexity. In other words, the network builds a hierarchical representation of the input, as the first layer represents the input in terms of elementary features and the deeper we go, the more abstract features can be recognized from the layers. The capture the complex features requires also to introduce some non-linearity in our system. Thus, non-linear functions are interjected between adjacent convolutional layers, and as a result a two-layer CNN can be proven to be a universal approximator \cite{hornik1991approximation}, while the identity function does not satisfy this property and generally, when multiple layers use the identity function, the entire network is equivalent to a single-layer model. Typical choices for the non-linear function, also known as activation function, include the logistic (sigmoid) function, the hyperbolic tangent (tanh), the Rectified Linear Unit (ReLU) and its variations \cite{xu2015empirical}.

Another important concept of CNNs is the pooling layer, which is a form of non-linear down-sampling. There are several non-linear functions to implement pooling, among which max pooling is the most common. Intuitively, the exact location of a feature is less important than its rough location relative to other features. This is the idea behind pooling in CNNs. The pooling layer progressively reduces the size of the representation, the number of parameters, the memory footprint and the amount of computation in the network, and hence also controls overfitting. It is common to periodically insert a pooling layer between successive convolutional layers in a CNN architecture.

Eventually, after several convolutional and max pooling layers, the high-level reasoning in the neural network is done via the Fully Connected layers (FC), commonly referred to as as dense layers. As its name implies, nodes in a fully connected layer have connections to all nodes in the previous layer leading to a very dense connectivity structure. Essentially, when the FC is inserted at the end of the architecture, it looks at the output of the previous layer, which represents the activation maps of high level features and determines which features are mostly correlated to a particular class. 

As a final classification step, we use the softmax activation function, which extends the idea of logistic regression into a multi-class world. That is, softmax assigns decimal probabilities to each class in a multi-class problem using the following equation:
\begin{eqnarray} \label{eq:1}
\sigma(x_i)=\frac{e^{x_i}}{\sum_{j=1}^{C} e^{x_j}} && \textnormal{for i=1,...,C}\
\end{eqnarray}
where $x$ is the input of the fully connected layer and $C$ is the total number of the distinct classes related to the problem at hand. This probabilistic approach renders possible to quantify the level of confidence for each estimation and provides a lucid view on what has been misconstrued in the case of misclassification.

\subsubsection{Recurrent Neural Networks}
Recurrent Neural Network (RNN) is a kind of neural network that specializes in processing sequences and has shown promising results in many Natural Language Processing (NLP) tasks \cite{ebrahimi2015chain}. The idea behind RNNs is the usage of sequential information, and they are called recurrent because they perform the same task for every element of a sequence, with the output being depended on the previous computations. Another way to think about RNNs is that they have a “memory” component, which captures information about what has been calculated thus far.

\begin{figure*}
\centering
\subfloat[]{\includegraphics[width=0.45\textwidth]{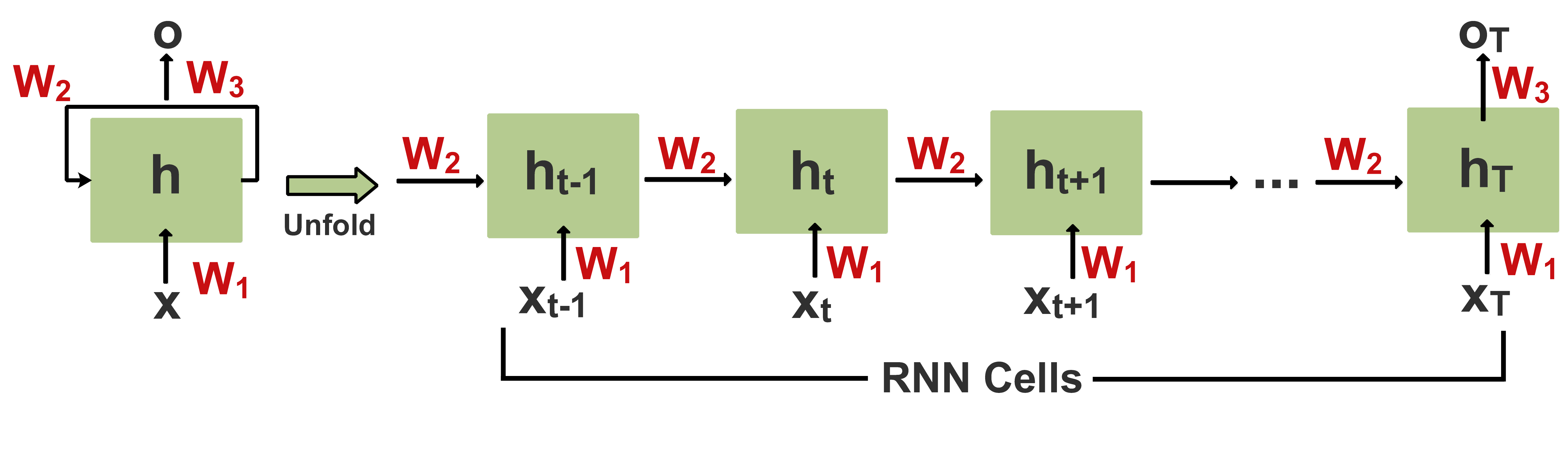}}
\hspace{10mm}
\subfloat[]{\includegraphics[width=0.45\textwidth]{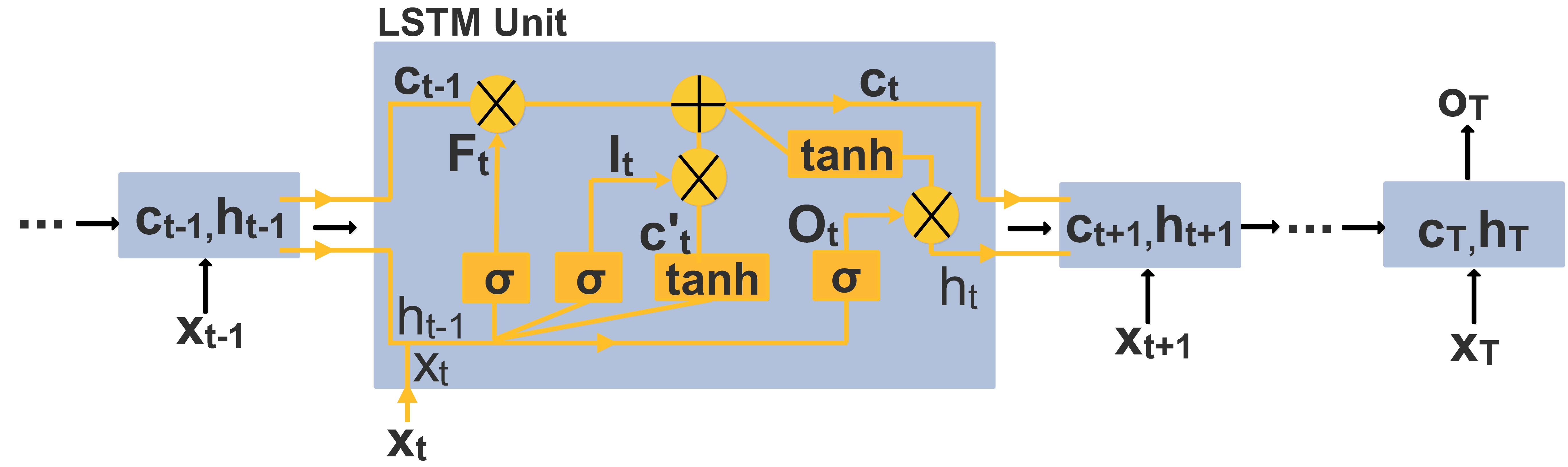}}\\
\subfloat[]{\includegraphics[width=0.8\textwidth]{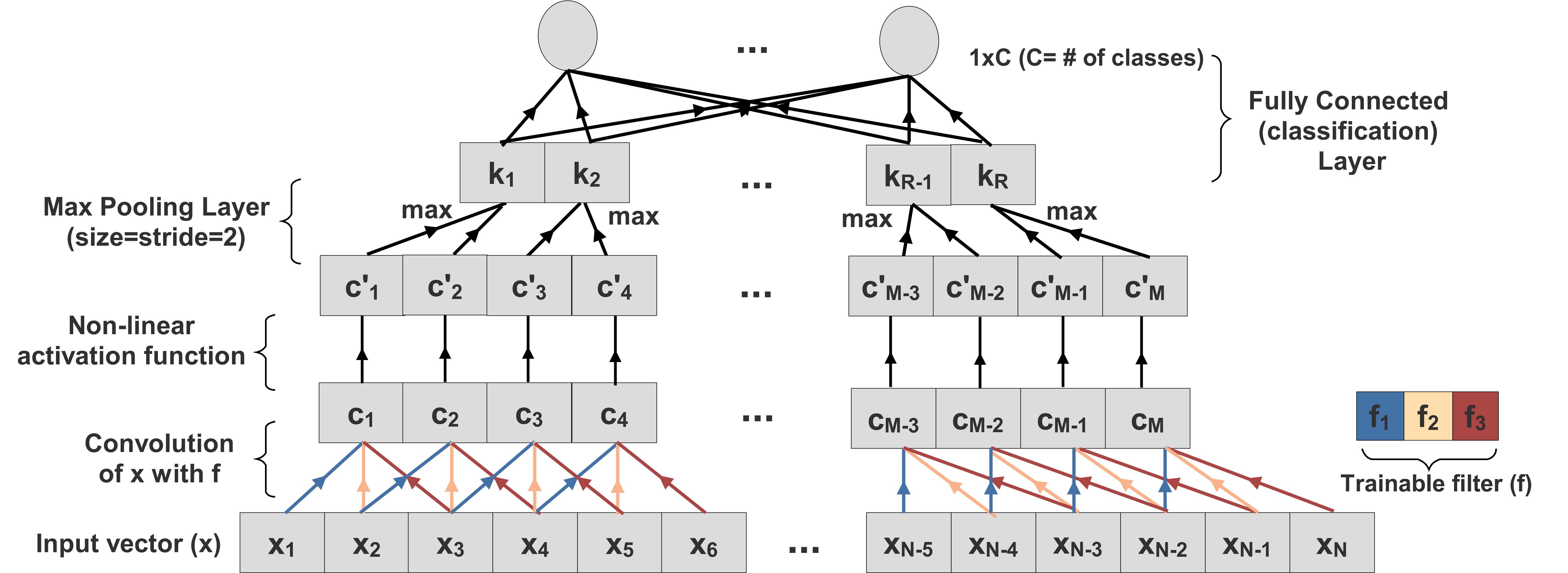}}
\caption{Proposed Deep Learning architectures for the neuronal cell-type classification: (a) RNN architecture: An RNN layer unfolded in $T$ RNN Cells, where $T$ is the total number of timesteps. Every cell receives at timestep $t$, the current value $x_t$ and the previous hidden state $h_{t-1}$ value as inputs and, in our case, only the last cell outputs a vector $o_T$, which represents the 4 distinct classes of our problem. (b) LSTM architecture: An LSTM layer unfolded in $T$ LSTM units. At timestep $t$ the gates $I$, $F$ and $O$ calculate their activations (i.e. $I_t$, $F_t$ and $O_t$ respectively) considering the current value $x_t$ and the activation of the memory cell at the previous timestep c\textsuperscript{t-1}. Circles containing the $X$ symbol represent an element-wise multiplication between its inputs. The rectangles containing a $\sigma$ symbol represent the application of the sigmoid differentiable function. At the final timestep $T$ the last LSTM unit outputs the vector $o_T$ with the 4 classes of the problem. (c) 1D-CNN architecture: The input vector $x$ is convolved with a trainable filter $f$ (stride equal to 1) resulting to a vector $c$, to which a non-linear activation function is applied, resulting to another vector $c'$ with the same size. A max pooling layer of size 2 is also applied to $c'$, in order to down-sample the input representation reducing its dimensionality. The number of the output nodes $C$ equals to the number of the classes (4 in our case). }
\label{fig_architectures}
\end{figure*}

The RNN model in our proposed method receives an input vector $x$ and gives an output vector $o$, which in our case are the input timeseries and the cell-type label, respectively. The diagram in Fig. \ref{fig_architectures}(a) shows an RNN architecture being unrolled (or unfolded) into a full network, which means that we write out the network for the complete sequence.

At timestep $t$, $x_t \in R^{1\times d}$ is the input entry, where $d$ is the dimensionality of that entry and $h_t$ is the hidden state at timestep $t$. Hidden state $h_t$ can be considered as the memory component of the network, given that it captures information about what happened in all previous timesteps. It is calculated based on the following equation and utilizes the previous hidden state as well as the input at the current timestep $t$: 
\begin{eqnarray} \label{eq:2}
h_t=f_1(W_1x_t+W_2h_{t-1}+b_h)
\end{eqnarray}
where function $f_1$ is usually non-linear, such as tanh or ReLU, $W_1$ and $W_2$ are the weight matrices used for both $x_t \rightarrow h_t$ and $h_{t-1} \rightarrow h_t$ links, respectively, whereas $b_h$ is the bias term added when calculating $h_t$. Hidden state $h_{-1}$, which is required for the calculation of the first hidden state, is typically initialized to zeroes. At the final timestep $T$, a dense layer calculates, as shown in Fig. \ref{fig_architectures}(a), the output $o_T$, given by the following equation:
\begin{eqnarray} \label{eq:3}
o_{T}=f_2(W_3h_{T}+b_o)
\end{eqnarray}
where $f_2$ is usually a softmax activation function (as in the case of 1D-CNNs, and generally when the task is a multi-class classification problem), which takes the logits of the dense layer and turns them into probabilities, and $b_o$ is the added bias. Note that depending on the task, Fig. \ref{fig_architectures}(a) could have an output $o_t$ at each timestep $t$, but for our application, which outputs a cell-type label given an input timeseries, only one output is calculated at the final timestep. Moreover, unlike a traditional DNN, which uses different parameters at each layer, an RNN model shares the same parameters ($W_1, W_2$), as depicted in Fig. \ref{fig_architectures}(a), across all timesteps. This reflects that we are performing the same task at each step, using only different inputs, which greatly reduces the total number of learnable parameters.

Despite the advantage of the RNNs to remember information through time, training an RNN is not a simple task with respect to the backpropagation algorithm, which is used as in the traditional neural networks but with a little twist. Because the parameters are shared by all timesteps in the network, the gradient at each output depends not only on the calculations of the current timestep, but also on the previous timesteps. For example, in order to calculate the gradient at $t=j$ we need to backpropagate $j-1$ steps and sum up the gradients. This is called Backpropagation Through Time (BPTT) and vanilla RNNs trained with BPTT have difficulties to learn long-term dependencies (i.e., dependencies between steps that are far away) \cite{pascanu2013difficulty}. This results in the so-called vanishing/exploding gradient problem. Thus, in order to address these issues, certain types of RNNs, such as LSTMs, which are described in the next subsection, were specifically designed to handle these drawbacks.

\subsubsection{Long Short-Term Memory Neural Networks}
Long Short-Term Memory is an artificial RNN model \cite{hochreiter1997long} used in the field of DL, which was developed to cope with the exploding/vanishing gradient problems that can be encountered when training traditional RNNs. An LSTM has a similar control flow as a RNN. It processes data passing on information as it propagates forward. The differences with an RNN layer are the operations within the LSTM's units. LSTMs, sometimes also combined with other methods, and various extensions or variants of LSTMs do not only process single data points, such as images \cite{byeon2015scene}, but also entire sequences of data, such as speech or video and therefore are applicable to tasks, such as speech \cite{graves2005framewise} and handwriting recognition \cite{graves2008unconstrained}, sign-language translation \cite{gao2017video}, etc. 

A typical LSTM architecture will contain several LSTM units (as many as the number of timesteps). A common architecture of an LSTM unit at timestep $t$, as shown in Fig. \ref{fig_architectures}(b), is composed of the cell state $c_t$ , which is the memory part of the LSTM unit, and three "regulators", usually called gates, that control the flow of information inside the LSTM unit. Specifically, a forget gate $F_t$, which decides what is relevant to be kept from prior steps, an input gate $I_t$, which decides what information is relevant to be added from the current step, and an output gate $O_t$, which determines what the next hidden state $h_t$ should be. Some variations \cite{xingjian2015convolutional} of the LSTM units do not include one or more of these gates, or they have other gates.

The first step of an LSTM unit is to decide what information is thrown away from the cell state. This decision is made by a sigmoid layer called the “forget gate layer”. Taken into account the $h_{t-1}$, which is the hidden state vector, commonly referred to as output vector of the LSTM unit, and the current information $x_t$, and outputs a number between $0$ and $1$ for each number in the cell state $c_{t-1}$. $1$ represents the “completely keep this information”, while a $0$ represents “completely get rid of this”. The equation for the forget gate layer is as follows:
\begin{eqnarray} \label{eq:4}
F_{t}=\sigma_g(W_Fx_{t}+R_Fh_{t-1}+b_F)
\end{eqnarray}
where $\sigma_g$ denotes the gate activation function, $W_F$, $R_F$ and $b_F$ are the learnable weights of an LSTM layer, i.e., the input, the recurrent weights and the bias for the forget layer component, respectively.

The next step consisting of two parts is to decide what new information will be stored in the cell state. First, a sigmoid layer, called the “input gate layer”, decides which values will be updated and a tanh layer creates a vector of new candidate values, $c'_t$ that could be added to the cell state. The equations describing these components at timestep $t$ are the following:
\begin{eqnarray} \label{eq:5}
I_{t}=\sigma_g(W_Ix_{t}+R_Ih_{t-1}+b_I)
\end{eqnarray}

\begin{eqnarray} \label{eq:6}
c'_{t}=\sigma_c(W_{c'}x_{t}+R_{c'}h_{t-1}+b_{c'})
\end{eqnarray}
where $\sigma_c$ denotes the candidate cell state tanh activation function and $W_I$, $R_I$, $b_I$ as well as $W_{c'}$, $R_{c'}$ and $b_{c'}$ are the learnable input and recurrent weights, and the bias for the input gate and cell candidate, respectively. These two steps are combined in order to update the old cell state $c_{t-1}$ into a new cell state $c_t$ based on the following equation:
\begin{eqnarray} \label{eq:7}
c_{t}=F_t*c_{t-1}+I_t*c'_t
\end{eqnarray}

Finally, the LSTM unit decides its output. The output id based on the cell state, but might be a filtered version. Firstly, a sigmoid layer, called the “output gate layer”, decides what parts of the cell state to output. Then, the cell state passes via a tanh, so that the values are re-scaled between $-1$ and $1$ and is multiplied with the output gate layer so that it outputs only the corresponding parts via the hidden state $h_t$. The equations describing these components are the following:
\begin{eqnarray} \label{eq:8}
O_{t}=\sigma_g(W_Ox_{t}+R_Oh_{t-1}+b_O)
\end{eqnarray}

\begin{eqnarray} \label{eq:9}
h_{t}=O_t*\sigma_c(c_t)
\end{eqnarray}
where $W_O$, $R_O$, $b_O$ in eq. \ref{eq:8} are the learnable input and recurrent weights and the bias for the output gate. Note that depending on the task, LSTMs could also output a label at each timestep $t$, but for our application, which outputs a cell-type label given an input timeseries, only one output is calculated at the final timestep.

\subsection{Regularization Methods}
Although DL architectures are very powerful machine learning systems, they contain a large number of parameters, which makes them quite complex models. As a DNN learns, weights settle into their context within the network. Weights of nodes are tuned for specific features providing some specialization. Nodes of neighboring layers rely on this specialization, which if taken too deep could result in a fragile model too specialized to the training data. This reliant on context for a node during training is referred to complex co-adaptations and can lead to overfitting of the training data, meaning that the network produces over-optimistic predictions throughout the training process, but fails to generalize well on new data leading to a significant drop in its performance.

Dropout \cite{srivastava2014dropout} is a regularization technique, which is essentially used to prevent overfitting while training ANNs and DNNs. The term “dropout” refers to dropping out units in a neural network, and thus these units are not considered during a particular forward and backward pass. More specifically, for the 1D-CNNs, at each training stage, individual nodes of a specific layer are either kept with probability \textit{p} or dropped out of the net with probability \textit{1-p}, so that a reduced network is left with the incoming and outgoing edges of the dropped-out node to be removed. Regarding the RNN and LSTM architectures, dropout can be also applied to the recurrent connections of these networks (i.e., the connections related to the hidden states), so that the recurrent weights could be regularized to improve performance. For any of the three architectures, each layer can be associated with a different probability \textit{p}, meaning that dropout can be considered as a per-layer operation with some layers discarding more nodes compared to others dropping nodes with a lower rate or no rate at all.

\subsection{Methodology}
Our 1D-CNN architecture accepts the input vector of Ca\textsuperscript{2+} signal $x \in R^{1\times 4000}$, which is standard normalized (z-score). The first convolutional layer of our system consists of 32 filters, while the rest of the layers consist of 62 filters, respectively. For the specific dataset studied, the most optimal depth (see next Section) is 3. Generally, the depth (i.e., number of convolutional layers) as well as the number of filters for the specific timeseries data application is kept small to avoid overfitting. Thus, regularization methods (i.e., dropout) are helpful for our system. Moreover, while in several applications, the size of the filter is 3 or 5, we used a trainable filter $f \in R^{1\times 10}$. Smaller filter sizes applied to our timeseries data are not effective enough, as Ca\textsuperscript{2+} signal has slow kinetics, and the signal is not notably distinguishable in less than 10 timesteps. In contrast, larger filter sizes lead to overfitting. Eventually, at the end of the network, a FC is attached, which receives the output of the previous layer and determines which features are mostly correlated to each one of the 4 classes.

Regarding RNN and LSTM models, they are usually more effective with timeseries data, whose length does not exceed the 100 timesteps in total, since they demand a considerable amount of time in order to be trained. Thus, in cases similar to ours, i.e. with long timeseries data, the following framework is proposed: each timeseries $X$ of length $N$ should be broken into $T$ timesteps with each timestep $x_t\in{R^d}$, where $d$ is the input dimensionality and $t=1,...,T$, such that $N=d \times T$ and $T\ll d$. Namely, each input timeseries $X$ will consist of $T$ timesteps, where each timestep is of dimensionality $d$. In our case, where each timeseries has a length of $N=4000$ timesteps, we break each timeseries in $T=2$, $T=5$ and $T=10$ time-steps, where each timestep $x_t$ is of dimensionality $d=2000$, $d=800$ and $d=400$, respectively.

\section{Experimental Analysis and Discussion}
\noindent
The DL models that were used in our analysis were implemented using the Tensorflow \cite{abadi2016tensorflow} and Keras open-source libraries written in Python programming language. 
Both TensorFlow and Keras can perform the calculations on GPUs, and thus the training time is dramatically reduced. For our experiments we used Python version $3.6$, the Tensorflow version $1.9$ running on NVIDIA GeForce GTX 750 Ti GPU model under Windows 10 operating system.

\subsection{Data Set}
\noindent The data set used to train the classifiers was collected during a goal oriented task in awake, behaving mice \cite{turi2019vasoactive}. Specifically, head-fixed mice ran on a 2-meter long treadmill belt equipped with a water delivery port (reward location) and the neural signals were recorded using the two-photon Ca\textsuperscript{2+} imaging technique. The mice learned the reward location after training on the belt for a few days. The recordings were obtained from the CA1 region of hippocampus, which is widely known to be involved in spatial memory formation. The data were then processed in order to translate the video recordings into fluorescence signals over time. Four different neuronal types were recorded during the aforementioned task. Namely, the excitatory PY cells, the PV, the SOM and the VIP inhibitory neurons making the problem a four-class classification task. Therefore, our design matrix consists of signals in time (timeseries) of four different neuronal types across all sessions/days and different animals.

\subsection{Impact of Regularization and Network's Depth}
In this subsection we study how the architecture of each model including the depth and dropout regularization hyper-parameters, affects the performance as well as the training time corresponding to 20 epochs. We present the results in Table \ref{First_Table}. We used $3947$ examples ($1000$ PY cells, $1000$ SOM cells, $1000$ VIP cells and $947$ PV cells), where each example is a timeseries that corresponds to a specific neuronal cell-type consisting of $4000$ timesteps (we have used the minimum length across all timeseries). We perform 10 random train-test splits, where in every split we use a fixed number of $3157$ training and $790$ testing examples, which have been z-score normalized based on the mean and standard deviation of the training set. Thus, we report the mean accuracy and training time of the 10 random train-test splits and their corresponding standard deviations. The hyper-parameter values for all the models have been selected after several pre-experiments in order to obtain the best set.

Regarding the architectures of the 1D-CNN model, which are presented in Table \ref{First_Table}, each convolutional layer is followed by a ReLU activation function and the pipeline ends up with a FC layer. The optimizer that is used in order to train the network is the Adam gradient-descent optimizer with learning rate $0.001$, and ${beta}_1$ as well as $beta_2$ parameters are $0.9$ and $0.99$ respectively. The first convolutional layer of each architecture is convolved with 32 filters with kernel size 10 and stride $1$, while as we go deeper, from the second convolutional layer and onwards, the inputs to succeeding layers are convolved with 64 filters of kernel size 10 and stride $1$. Table \ref{First_Table} shows that the optimal architecture is the fifth one (bolded mean accuracy), which is composed of 2 convolutional layers followed by a max pooling layer of size and stride equal to $2$ followed by a last convolutional layer and a dropout layer.

\begin{table*}[h]
\scriptsize
\renewcommand{\arraystretch}{1.6}
\centering
\begin{tabular}{|c|c||c|c|c|c|c|c|}
\hline
\textbf{Models} & \textbf{Depth} & \textbf{Mean Acc.} & \textbf{St. Dev.} & \textbf{Mean Tr. Time (sec.)} & \textbf{St. Dev.} \\ \hhline{|=|=|=|=|=|=|}
\multirow{8}{*}{\rotatebox{90}{1D-CNN}} & \begin{tabular}[c]{@{}c@{}}1 Conv. Layer\end{tabular} & $0.8368$ & $0.0116$ & $50.2281$ & $0.8775$ \\ \cline{2-8} 
& 2 Conv. Layers & $0.8674$ & $0.0116$ & $124.0125$ & $2.7903$ \\ \cline{2-8} 
& 2 Conv. Layers-Max P.-1 Conv. Layer & $0.8739$ & $0.0125$ & $155.3015$ & $1.9314$ \\  \cline{2-8}
& 2 Conv. Layers-Max P.-2 Conv. Layer & $0.8746$ & $0.0102$ & $188.139$ & $2.9593$ \\ \cline{2-8} 
& 2 Conv. Layers-Max P.-1 Conv. Layer-Dropout & $\bf{0.8867}$ & $0.0119$ & $161.9093$ & $2.3241$ \\ \cline{2-8}
& 2 Conv. Layers-Max P.-1 Conv. Layer-Dropout-1 Conv. Layer & $0.8683$ & $0.0114$ & $197.5203$ & $3.5911$ \\ \cline{2-8}
& 2 Conv. Layers-Dropout-2 Conv. Layers & $0.8394$ & $0.0219$ & $345.664$ & $2.3541$ \\ \cline{2-8}
& 3 Conv. Layers & $0.8502$ & $0.032$ & $226.925$ & $2.5193$ \\ \hhline{|=|=|=|=|=|=|}
\multirow{3}{*}{\rotatebox{90}{RNN}} \multirow{3}{*}{\rotatebox{90}{2 timesteps}}  &  \begin{tabular}[c]{@{}c@{}} 1 RNN Layer\end{tabular} & $\bf{0.8205}$ & $0.011$ & $38.6984$ & $0.6726$\\ \cline{2-8} 
& 2 RNN Layers & $0.804$ & $0.03$ & $55.1843$ & $1.1724$ \\ \cline{2-8} & 2 RNN Layers-Dropout-1 RNN Layer & $0.8067$ & $0.0141$ & $71.5421$ & $0.9292$\\ \hhline{|=|=|=|=|=|=|}
\multirow{3}{*}{\rotatebox{90}{RNN}} \multirow{3}{*}{\rotatebox{90}{5 timesteps}}  & \begin{tabular}[c]{@{}c@{}}1 RNN Layer\end{tabular} & $0.8167$ & $0.01627$ & $49.925$ & $0.6941$\\ \cline{2-8} 
& 2 RNN Layers & $0.8065$ & $0.0151$ & $80.2203$ & $1.1572$ \\ \cline{2-8} & 2 RNN Layers-Dropout-1 RNN Layer & $0.7975$ & $0.0227$ & $109.6156$ & $1.2772$\\ \hhline{|=|=|=|=|=|=|}
\multirow{3}{*}{\rotatebox{90}{RNN}} \multirow{3}{*}{\rotatebox{90}{10 timesteps}}  & \begin{tabular}[c]{@{}c@{}}1 RNN Layer\end{tabular} & $0.8026$ & $0.0194$ & $71.8468$ & $0.7897$\\ \cline{2-8} 
& 2 RNN Layers & $0.8024$ & $0.0273$ & $124.6187$ & $0.979$ \\ \cline{2-8} & 2 RNN Layers-Dropout-1 RNN Layer & $0.7969$ & $0.0133$ & $170.7687$ & $1.6552$\\ \hhline{|=|=|=|=|=|=|}
\multirow{3}{*}{\rotatebox{90}{LSTM}} \multirow{3}{*}{\rotatebox{90}{2 timesteps}} & \begin{tabular}[c]{@{}c@{}}1 LSTM Layer\end{tabular} & $0.7734$ & $0.0166$ & $69.35$ & $0.8836$ \\ \cline{2-8} 
&  2 LSTM Layers & $\bf{0.7915}$ & $0.015$ & $102.8062$ & $1.7754$ \\ \cline{2-8} 
& 2 LSTM Layers-Dropout-1 LSTM Layer & $0.7897$ & $0.01$ & $141.7281$ & $2.1877$ \\ \hhline{|=|=|=|=|=|=|}
\multirow{3}{*}{\rotatebox{90}{LSTM}} \multirow{3}{*}{\rotatebox{90}{5 timesteps}} & \begin{tabular}[c]{@{}c@{}}1 LSTM Layer\end{tabular} & $0.7869$ & $0.0211$ & $91.5953$ & $1.3077$ \\ \cline{2-8} 
& 2 LSTM Layers & $0.7822$ & $0.0158$ & $156.3843$ & $2.2983$ \\ \cline{2-8} 
& 2 LSTM Layers-Dropout-1 LSTM Layer & $0.7648$ & $0.018$ & $228.5359$ & $3.7028$ \\ \hhline{|=|=|=|=|=|=|}
\multirow{3}{*}{\rotatebox{90}{LSTM}} \multirow{3}{*}{\rotatebox{90}{10 timesteps}} & \begin{tabular}[c]{@{}c@{}}1 LSTM Layer\end{tabular} & $0.7911$ & $0.0111$ & $140.9765$ & $1.3509$ \\ \cline{2-8} 
& 2 LSTM Layers & $0.7896$ & $0.0171$ & $256.5265$ & $1.333$ \\ \cline{2-8} 
& 2 LSTM Layers-Dropout-1 LSTM Layer & $0.6464$ & $0.1118$ & $381.5828$ & $4.6384$ \\ \hhline{|=|=|=|=|=|=|=|=|}  
\end{tabular}\\[5pt]
\caption{Mean accuracy performance of the DL architectures on neuronal cell-type classification and their corresponding mean training time across 10 random train-test splits.}
\label{First_Table}
\end{table*}
We observe that the architecture consisting of $2$ convolutional layers gives a better classification performance compared to the architectures consisting of $1$ and $3$ convolutional layers, as by using only $1$ layer, we create a very shallow network, which cannot be trained properly, while $3$ layers lead to overfitting. By using the dropout regularization technique combined with a max pooling layer in order to control overfitting, we observe that the $5^{th}$ architecture, where the dropout layer is inserted just before the FC layer is the most effective one, as FC layers are more prone to overfitting due to their large number of connections. Training time, as expected, increases by adding more convolutional layers.  

Regarding the RNN and LSTM models, each RNN and LSTM layer is followed by a ReLU activation function and the pipeline ends up with a FC layer. Moreover, each RNN and LSTM layer consists of $100$ hidden units, which is essentially the dimensionality of the output space. The optimizer that we used in order to train the network is the Adam gradient-descent based algorithm with the same parameters as before.

More specifically, the optimal architecture for the RNN model, as shown in Table \ref{First_Table} is obtained in the case of $2$ timesteps for a single RNN layer. Moreover, for all different values of timesteps, the single RNN layer architecture gives better classification results compared to the stacked RNNs (i.e., an RNN with more than one layer), which lead to network overfitting. In order to prevent overfitting, we added a dropout layer, but the performance was not improved (Table \ref{First_Table}). We also experimented by adding recurrent dropout layers with and without the dropout layer but the accuracy performance was significantly dropped (i.e., $1\%-3\%$ lower, data not shown). We observe again that increasing the complexity of the architecture by adding extra RNN layers, the training time is also increased. Regarding the LSTM model, the highest performing architecture is obtained in the case of $2$ timesteps with $2$ stacked LSTM layers (Table \ref{First_Table}). In general, similarly to RNN model, various combinations of dropout and recurrent dropout layers among the LSTM layers did not improve the classification performance.
\begin{figure}[h]
\centering
\subfloat[1D-CNN]{\includegraphics[width=0.24\textwidth]{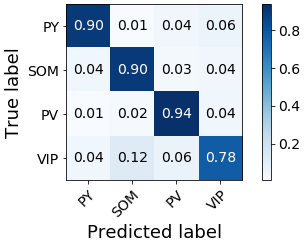}}
\subfloat[RNN]{\includegraphics[width=0.24\textwidth]{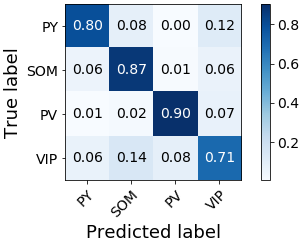}}\\
\subfloat[LSTM]{\includegraphics[width=0.24\textwidth]{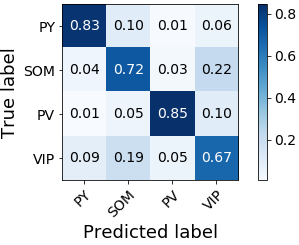}}
\caption{Normalized confusion matrices for the best-performing 1D-CNN, RNN and LSTM models.}
\label{fig_cm}
\end{figure}

Fig. \ref{fig_cm} demonstrates the normalized confusion matrices corresponding to the optimal architectures of the three models, as reported in Table \ref{First_Table} (given in bolded). We observe that for all models, classification errors concern primarily the VIP cell type, while the PY and PV cells are identified with higher accuracy.

We observe that 1D-CNN is the optimal model for this task (Table \ref{First_Table} and Fig. \ref{fig_cm}), as it generates the most accurate predictions. This finding suggests that long-term dependencies are unlikely to be significant in the particular scenario. This is because both RNN and LSTM models, which are capable of identifying such dependencies, have a poorer performance than the CNN architecture. The efficacy of 1D-CNN can be attributed to the fact that they take bi-directional temporal information into account, compared to the single-temporal-direction features extracted by RNN and LSTM models.

We also applied our proposed scheme with the data re-organization when training the RNN and LSTM models, and we observed that as we reduce the number of timesteps $T$, and hence automatically increase the input dimensionality $d$, the classification accuracy is improved for both models, and the mean training time gets also substantially decreased (Table \ref{First_Table}). We also examined the case of $T=4000$ timesteps and $d=1$, which practically means that we feed the network with a scalar value at each timestep. Performance was very poor, as the maximum classification accuracy that was achieved was around $45\%$ for both models, and the mean training time was dramatically increased especially for the LSTM model. Thus, we conclude that for the specific task when using these two models, a higher input dimensionality is more effective than an unfolding of many timesteps, which strengthens again the aforementioned finding that no significant long-term dependencies exist in the input data. 

Overall, we observe that the LSTM model has a lower performance compared to the RNN, with respect to both the classification accuracy and the training time. This could be attributed to overfitting, as the LSTMs are structurally more complex models compared to RNNs, and  given the fact that we did not use large amounts of training data, this intrinsic complexity of the LSTM model is probably unnecessary for the modeling of the specific dataset. Overfitting is also evidenced by the achieved training accuracy during the last epochs. While 1D-CNN and RNN models reached a training accuracy of $95\%-100\%$, the LSTM model achieved a lower training accuracy of $88\%-97\%$.

We also investigated the impact of the number of training epochs in the accuracy of the tested architectures. Specifically, we trained our best-performing models for $60$, and $100$ training epochs and compared their performance with the one obtained when $20$ epochs are used. For the optimal 1D-CNN architecture, the resulting accuracy was $0.8787$ and $0.881$, for $60$ and $100$ epochs respectively, compared to $0.8867$ for $20$ epochs. This demonstrates that the number of $20$ epochs chosen to train the 1D-CNN model is adequate. Similarly, for the best-performing RNN model, the accuracies achieved were $0.8188$ and $0.8169$ for $60$ and $100$ epochs, respectively compared to $0.8205$ for $20$ epochs. Finally, for the best-performing LSTM model, the accuracies achieved were $0.7911$ and $0.7894$ for $60$ and $100$ epochs, respectively, compared to $0.7915$ for $20$ epochs. Overall, this analysis confirmed that using $20$ epochs is sufficient for ensuring the optimal performance while avoiding overfitting and larger amount of training time.

Table \ref{metrics_Table} shows the precision, recall and specificity metrics (for each neuronal cell-type separately) derived from our best-performing model (i.e., the 1D-CNN architecture, whose mean accuracy in Table \ref{First_Table} is in bold). 

\begin{table}[h]
\renewcommand{\arraystretch}{1.4}
\centering
\begin{tabular}{cccc}
\\ \hline
\multicolumn{1}{|c||}{\bf{Metrics}} & \multicolumn{1}{c|}{\bf{precision}} & \multicolumn{1}{c|}{\bf{recall}}    & \multicolumn{1}{c|}{\bf{specificity}}   \\ \hline
\multicolumn{1}{|c||}{PY}      & \multicolumn{1}{c|}{0.908 (0.05)} & \multicolumn{1}{c|}{0.885 (0.06)} & \multicolumn{1}{c|}{0.967 (0.02)} \\ \hline
\multicolumn{1}{|c||}{SOM}      & \multicolumn{1}{c|}{0.874 (0.05)} & \multicolumn{1}{c|}{0.913 (0.03)} & \multicolumn{1}{c|}{0.954 (0.02)} \\ \hline
\multicolumn{1}{|c||}{PV}      & \multicolumn{1}{c|}{0.932 (0.01)} & \multicolumn{1}{c|}{0.916 (0.02)} & \multicolumn{1}{c|}{0.977 (0.006)} \\ \hline
\multicolumn{1}{|c||}{VIP}      & \multicolumn{1}{c|}{0.819 (0.06)} & \multicolumn{1}{c|}{0.779 (0.05)} & \multicolumn{1}{c|}{0.938 (0.03)} \\ \hline
\end{tabular}\\[3pt]
\caption{Mean performance and standard deviation (parentheses) of the best-performing 1D-CNN model for each cell-type}
\label{metrics_Table}
\end{table}

\subsection{Impact of the training set size}
Unlike other types of datasets used in machine learning like Imagenet, whose size is directly related to the amount of human effort involved in annotating images, the case of biological data offers a significantly more challenging setting. This is due to the fact that annotations, cell types in this case, are obtained though the postmortem biophysical analysis of cells. As such, the availability of training/validation data is substantially more limited compared to other cases. One of the objectives in this paper is also to understand the performance of different DL methods given a relativity limited set of training examples. 

Table \ref{Forth_Table} demonstrates how the size of the training set affects the mean accuracy of the best-performing 1D-CNN, RNN and LSTM models (i.e., models whose mean accuracy in Table \ref{First_Table} is in bold). We used 4 different training set sizes, but in each case we retained a fixed-size testing set of $640$ examples for a fair comparison.

For a training set of $2560$, we used $640$ cells from each category, while for a training set of $3157$ we used $800$ examples from each category of PY, SOM and VIP, and $757$ PV cells. For the dataset with $5137$ training examples we used $1600$ examples from each of the PY and VIP classes, $1180$ SOM cells and $757$ PV cells. Finally, in the case of $6737$ training examples we used $2400$ examples from each of the PY and VIP classes, $1180$ SOM cells and $757$ PV cells. 

We found that differences in the performance across different training set sizes were small for all types of models. Specifically, the largest difference was observed between the 2 extreme cases of $2560$ and $6737$ training examples and was smaller than $6-7$\%. However, this relatively small improvement came with a very high cost of required time for training the models. Indicatively, the mean training times required by the 1D-CNN, given the extreme cases of $2560$ and $6737$ examples were $132.46$ and $361.155$ seconds, respectively. The same observation is derived from the other two models, where RNN needs $20.078$ and $51.643$ seconds and the LSTM $52.257$ and $129.973$ seconds.

\begin{table}[h]
\renewcommand{\arraystretch}{1.4}
\centering
\begin{tabular}{cccc}
\\ \hline
\multicolumn{1}{|c||}{\bf{Training Set Size}} & \multicolumn{1}{c|}{\bf{1D-CNN}} & \multicolumn{1}{c|}{\bf{RNN}}    & \multicolumn{1}{c|}{\bf{LSTM}}   \\ \hline
\multicolumn{1}{|c||}{2560}      & \multicolumn{1}{c|}{0.854 (0.02)} & \multicolumn{1}{c|}{0.792 (0.01)} & \multicolumn{1}{c|}{0.794 (0.01)} \\ \hline
\multicolumn{1}{|c||}{3157}      & \multicolumn{1}{c|}{0.88 (0.01)} & \multicolumn{1}{c|}{0.829 (0.01)} & \multicolumn{1}{c|}{0.803 (0.01)} \\ \hline
\multicolumn{1}{|c||}{5137}      & \multicolumn{1}{c|}{0.907 (0.008)} & \multicolumn{1}{c|}{0.833 (0.01)} & \multicolumn{1}{c|}{0.813 (0.01)} \\ \hline
\multicolumn{1}{|c||}{6737}      & \multicolumn{1}{c|}{0.918 (0.01)} & \multicolumn{1}{c|}{0.868 (0.01)} & \multicolumn{1}{c|}{0.848 (0.009)} \\ \hline
\end{tabular}\\[3pt]
\caption{Mean accuracy and standard deviation (parentheses) of the best-performing models for training sets of different size.}
\label{Forth_Table}
\end{table}

\subsection{Comparison with other classifiers}
To assess whether our approach is better than other types of popular classifiers, we compared the best-performing 1D-CNN, RNN and LSTM models (models of Table \ref{First_Table} in bold), against the traditional Machine Learning approaches, such as the Adaboost and the Support Vector Machine (SVM) classifiers. In particular, we used SVMs with linear, Gaussian and polynomial kernels. Table \ref{Second_Table} corroborates the claim that 1D-CNN is the most efficient algorithm for the task at hand. Its main competitor is the Gaussian SVM, which outperforms all other models except the RNN, while Adaboost \cite{scikit-learn} has the worst performance.

\begin{table}[]
\renewcommand{\arraystretch}{1.2}
\centering
\begin{tabular}{|c||c|c|c|c|}
\hline
\textbf{Model}       & \textbf{Mean Accuracy} & \textbf{St. Dev.} \\ \hline \hline
1D-CNN 		& $\bf{0.8867}$  & $0.0119$   \\ \hline
RNN 	    & $0.8205$ & $0.011$ \\ \hline
LSTM 	    & $0.7915$ & $0.015$ \\ \hline
Linear SVM 	     & $0.6168$ & $0.0125$ \\ \hline
Gaussian SVM 	 & $0.8184$ & $0.0108$ \\ \hline
Polynomial SVM	 & $0.6477$ & $0.0152$ \\ \hline
Adaboost	     & $0.568$ & $0.01$ \\ \hline
\end{tabular}\\[3pt]
\caption{Comparison of the best-performing 1D-CNN, RNN and LSTM models against the Adaboost and SVM classifiers.}
\label{Second_Table}
\end{table}

In order to infer the most efficient combination of parameter values for each of the classifiers, we used the GridSearchCV implementation \cite{scikit-learn}, which performs an exhaustive search over the hyperparameter space. The specific hyperparameter related to the Adaboost classifier is the number of estimators at which boosting is terminated. Correspondingly, the hyperparameter related to the Gaussian and polynomial SVMs are the gamma and the penalty parameters. For the polynomial SVM, the degree hyperparameter is also defined, while the hyperparameter related to the linear SVM classifier is the penalty parameter. Gamma is a hyperparameter for the non-linear hyperplanes. The higher its value the harder the classifier tries to fit the training data set. The penalty parameter of the error term controls the trade off between a smooth decision boundary and classifying the training points correctly. Degree is a parameter used with a polynomial kernel and is essentially the degree of the polynomial kernel used to find the hyperplane to split the data.

Using GridSearchCV, we found that the Adaboost classifier works best with $100$ estimators (i.e., classifiers). Regarding the Gaussian SVM, gamma equal to $0.0001$ and the penalty parameter equal to $100$ are the best-performing combinations. For the polynomial SVM the optimal combination of hyperparameter is a polynomial degree equal to $3$, and the gamma as well as the penalty parameters are equal to $0.001$ and $100$, respectively. Eventually, for the linear classifier, the most optimal penalty parameter equals to $10$. In general, pre-experiments showed that increasing the values of gamma and the penalty parameters leads to overfitting, as the classifier tries to perfectly fit the training data, while for smaller values, the classifier is not trained properly making more errors during the testing phase.

\begin{figure}[h]
\centering
\includegraphics[width=0.48\textwidth]{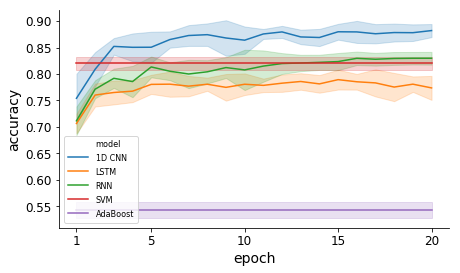}
\caption{Comparison of 1D-CNN, RNN and LSTM with traditional machine learning models (Gaussian SVM, AdaBoost). Solid lines denote means, while shaded areas standard deviation across 10 random train-test splits respectively.}
\label{fig_accuracies}
\end{figure}
Fig. \ref{fig_accuracies} compares the performance of the proposed DL models with traditional machine learning methods, such as the Gaussian SVM and Adaboost. Solid lines denote means, while the shaded areas denote standard deviation for the 10 random train-test splits. We observe that the mean performance of the DL models improves across epochs (having also small standard deviations), until they reach a plateau at $20$ epochs or earlier. Thus, while 1D-CNN starts in the first epoch with almost $75$\% accuracy, it eventually outperforms all other models. Adaboost on the other hand, has the worst performance.

\subsection{Impact of the number of timesteps}
\noindent 
In this subsection, we assess how the classification accuracy and the training time of the best-performing architectures (Table \ref{First_Table}) are affected by the number of timesteps in the input. As expected, the results in Table \ref{Third_Table} confirm that decreasing the number of timesteps in the input, results in lower performance accuracy and smaller training time. However, when $2000$ timesteps are used, while the classification accuracy (especially for the 1D-CNN model) has a small drop compared to the $4000$ timesteps, the training time has a sharp decrease, suggesting that the use of $2000$ timesteps is an acceptable option for fast and accurate cell-type classification. 

\begin{table}[h]
\renewcommand{\arraystretch}{1.4}
\centering
\begin{tabular}{cccc}
\multicolumn{1}{l}{}    & \multicolumn{3}{c}{\textbf{Mean Acc. \& Time (sec.)}}                                                       \\ \hline
\multicolumn{1}{|c||}{\textbf{Timesteps}} & \multicolumn{1}{c|}{\bf{1D-CNN}} & \multicolumn{1}{c|}{\bf{RNN}}    & \multicolumn{1}{c|}{\bf{LSTM}}   \\ \hline
\multicolumn{1}{|c||}{1000}      & \multicolumn{1}{c|}{0.8097 (61.314)} & \multicolumn{1}{c|}{0.7184 (35.793)} & \multicolumn{1}{c|}{0.7246 (90.784)} \\ \hline
\multicolumn{1}{|c||}{2000}      & \multicolumn{1}{c|}{0.8505 (96.140)} & \multicolumn{1}{c|}{0.7715 (37.95)} & \multicolumn{1}{c|}{0.7651 (94.021)} \\ \hline
\multicolumn{1}{|c||}{4000}      & \multicolumn{1}{c|}{0.8867 (161.90)} & \multicolumn{1}{c|}{0.8205 (38.698)} & \multicolumn{1}{c|}{0.7915 (102.806)} \\ \hline
\end{tabular}\\[3pt]
\caption{Mean accuracy and training time of the best-performing methods for various numbers of timesteps}
\label{Third_Table}
\end{table}

\subsection{Testing the models on a new dataset}
To assess the generalization performance of our models, we used a completely new dataset, derived from  different animals performing a different behavioral task (removal of the reward location, random foraging, unpublished results from the Losonczy lab). The new dataset includes $119$ timeseries of length $2606$ timesteps, where $91$ of the timeseries corresponded to the activity of SOM interneurons, while the rest of them corresponded to the activity of the PV interneurons. 

To ensure equal-length examples, we reduced the length of examples in the first dataset from $4000$ to $2606$ timesteps. We then re-trained the best-performing models using only the first dataset and tested them on both datasets (i.e., the $210$ new examples were added to the $790$ testing examples considered in the previous sections). The mean accuracy that was achieved by the 1D-CNN model was $81.65\%$, while the RNN and LSTM models achieved a mean accuracy of $78.3\%$ and $73.28\%$, respectively. These results are quite promising given that the datasets are completely different and the task performed by the animals is not identical. They further suggest that discriminatory features of the different cell types may be behavior-independent, thus making our approach an extremely powerful tool for the online classification of cell types in behaving animals.

\begin{figure}[h]
\centering
\subfloat[1D-CNN]{\includegraphics[width=0.24\textwidth]{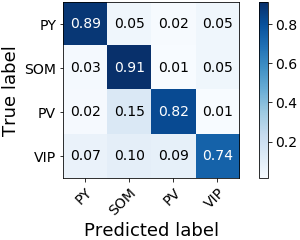}}
\subfloat[RNN]{\includegraphics[width=0.24\textwidth]{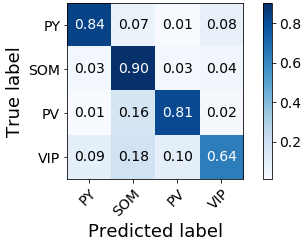}}\\
\subfloat[LSTM]{\includegraphics[width=0.24\textwidth]{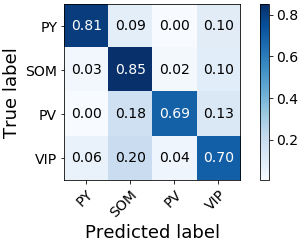}}
\caption{Normalized confusion matrices of the best-performing architectures of 1D-CNN, RNN and LSTM models tested on a new dataset}
\label{fig_cm_new_data}
\end{figure}

As shown in Fig. \ref{fig_cm_new_data}, all models misclassify more often the VIP and PV neurons, which are mainly predicted to be SOM cells. More specifically, the VIP neurons are again misclassified as SOM cells, as in the first dataset (Fig. \ref{fig_cm}), while now, PV neurons are also misclassified as SOM cells. This could be explained by the small number of the newly added PV cells (just $28$), which probably hinders their correct recognition by the classifier, compared to the $191$ newly added SOM cells, where there is more information to be exploited.

\section{Conclusion}
\noindent In this work we propose a DL-based formalization for the task of automatic neuronal cell-type classification based on the Ca\textsuperscript{2+} activity signal of neurons in behaving animals. We considered 1D-CNN, RNN and LSTM architectures and showed that these DL network architectures are capable of capturing hidden dynamics/features in the activity signal and therefore to make accurate predictions. We found that 1D-CNN is the optimal classifier for this task, suggesting the absence of significant long-term dependencies in the particular dataset. Our results reveal a great potential for replacing current approaches for cell-type classification (e.g., with molecular markers and/or feature extraction algorithms) with 1D-CNN Ca\textsuperscript{2+} imaging-based classifiers, making neuronal cell-type classification automate. 

Moreover, our results set the stage for a deeper investigation of whether automated classification of neuronal subtypes (i.e., basket and axoaxonic cells that are subtypes of of the PV cells) is possible. Our current datasets consist of large families of these subtypes, whose automated discrimination, if possible, will have major contribution to the design of experiments, leading to important resource savings (cost, time, effort and number of animals). Another avenue of application involves the development of a continuous learning ANN. This network should be able to understand when new cell-types are introduced during testing, and adjust its parameters so as to recognize them in a next appearance without requiring to re-train its parameters. These two research perspectives will be investigated in future work.


%



\ifCLASSOPTIONcompsoc
  \section*{Acknowledgments}
\else
  \section*{Acknowledgment}
\fi
\noindent
This research is co-funded by Greece and the European Union (European Social Fund-ESF) through the Operational Programme 'Human Resources Development, Education and Lifelong Learning' in the context of the project 'Strengthening Human Resources Research Potential via Doctorate Research' (MIS-5000432), implemented by the State Scholarships
Foundation (IKY).

\ifCLASSOPTIONcaptionsoff
  \newpage
\fi



%




\bibliographystyle{IEEEtran}

\bibliography{refs}

\end{document}